\newcommand{\rka}[1]{\cellcolor{red!28}#1}     % 第一名 light red
\newcommand{\rkb}[1]{\cellcolor{orange!28}#1}  % 第二名 light orange
\newcommand{\rkc}[1]{\cellcolor{yellow!35}#1}  % 第三名 light yellow
\newcommand{\sgain}[1]{\textcolor[rgb]{0.00,0.45,0.25}{(#1)}}
\newcommand{\sloss}[1]{\textcolor{red}{(#1)}}
\documentclass[]{bytedance_seed}
\usepackage[table,dvipsnames]{xcolor}
\usepackage{makecell}  
\usepackage{minitoc}
\usepackage{url}
\usepackage{amssymb}
\usepackage[utf8]{inputenc}
\usepackage{microtype}
\usepackage{booktabs}
\usepackage{pifont} 
\usepackage{multirow}
\usepackage{makecell}
\usepackage{paralist}
\usepackage{xspace}
\usepackage{color}
\usepackage{xcolor}
\usepackage{colortbl}
\usepackage{adjustbox}
\usepackage{hyperref} 
\usepackage[edges]{forest}
\usepackage{tikz} 
\usepackage{caption}
\usepackage{amsfonts}
\usepackage[toc,page,header]{appendix}
\usepackage[utf8]{inputenc}
\usepackage{bbm}
\usepackage{enumitem}
\usepackage{booktabs}
\definecolor{seedc}{RGB}{7, 92, 173}
\usepackage{cleveref}
\usepackage{fancyvrb}
\usepackage{CJKutf8}
\usepackage{threeparttable}

\newcommand{\name}[1]{Robix-2}

\renewcommand{\paragraph}[1]{\vspace{0.1em}\noindent\textbf{#1}}

\usepackage[most]{tcolorbox} % 加载 tcolorbox 宏包及其大部分库
\usepackage{xcolor}          % 用于定义颜色
\usepackage{lipsum}          % 用于生成示例长文本
\definecolor{usercolor}{rgb}{0.88, 0.96, 1.0}  % 淡蓝色
\definecolor{robotcolor}{rgb}{0.95, 0.95, 0.95} % 淡灰色

\newlength{\dialogboxthickness}
\title{GST-Bench: Can VLMs Develop Global Spatial Awareness from Video?}

\author[1,2,*]{Qifeng Zhang}
\author[2]{Kaixiang Huang}
\author[1]{Heng Dong} 
\author[1]{Huang Fang}
\author[1,3,*]{Junting Chen}
\author[2]{Junjie Zhu}
\author[2]{Yonghang Chen}
\author[2]{Zhiyu Zhang}
\author[1, \dagger]{\protect Wei Li}
\affiliation[1]{ByteDance Seed}
\affiliation[2]{Zhejiang University}
\affiliation[3]{National University of Singapore}

\contribution[*]{Work done at ByteDance Seed}
\contribution[\dagger]{Corresponding authors}

\abstract{
Spatial intelligence is fundamental to embodied agents, yet existing benchmarks focus on local spatial perception from single or few viewpoints, overlooking global spatial awareness over continuous, long-horizon visual streams. To address this limitation, we introduce the Global-Spatial-Temporal Benchmark (\textbf{GST-Bench}), a VQA benchmark for global spatial intelligence in video understanding, comprising human-verified questions derived from 6,790 minutes of synthetically generated video. It requires models to perform accurate spatial inference from novel viewpoints unseen in the input video and to map egocentric observations onto global top-down images. 
A comprehensive evaluation of 22 state-of-the-art VLMs exposes a striking gap between models and humans: the strongest zero-shot model attains only 42.68, far below the human score of 79.08. 
To probe the cause of this gap, we construct \textbf{GST-Bench-Local} and find that models, despite strong local spatial understanding under the same task formulation, still fail to consolidate long-horizon observations into a globally consistent scene representation.
We further provide \textbf{GST-Train}, a dataset for global spatial reasoning, as a complementary resource to facilitate future research on this challenge.
}

\date{\today}
\correspondence{\email{liwei.85@bytedance.com}}
\checkdata[Project Page]{\url{https://qwerirwq.github.io/GST-Bench/}}

\begin{document}
\maketitle

%不需要目录就注释掉 注意目录不要和第一页放在一块 要有\newpage
%\newpage
%\tableofcontents
%\newpage

\section{Introduction}
\label{sect:intro}

\begin{figure*}[t]
\centering
\includegraphics[width=\textwidth]{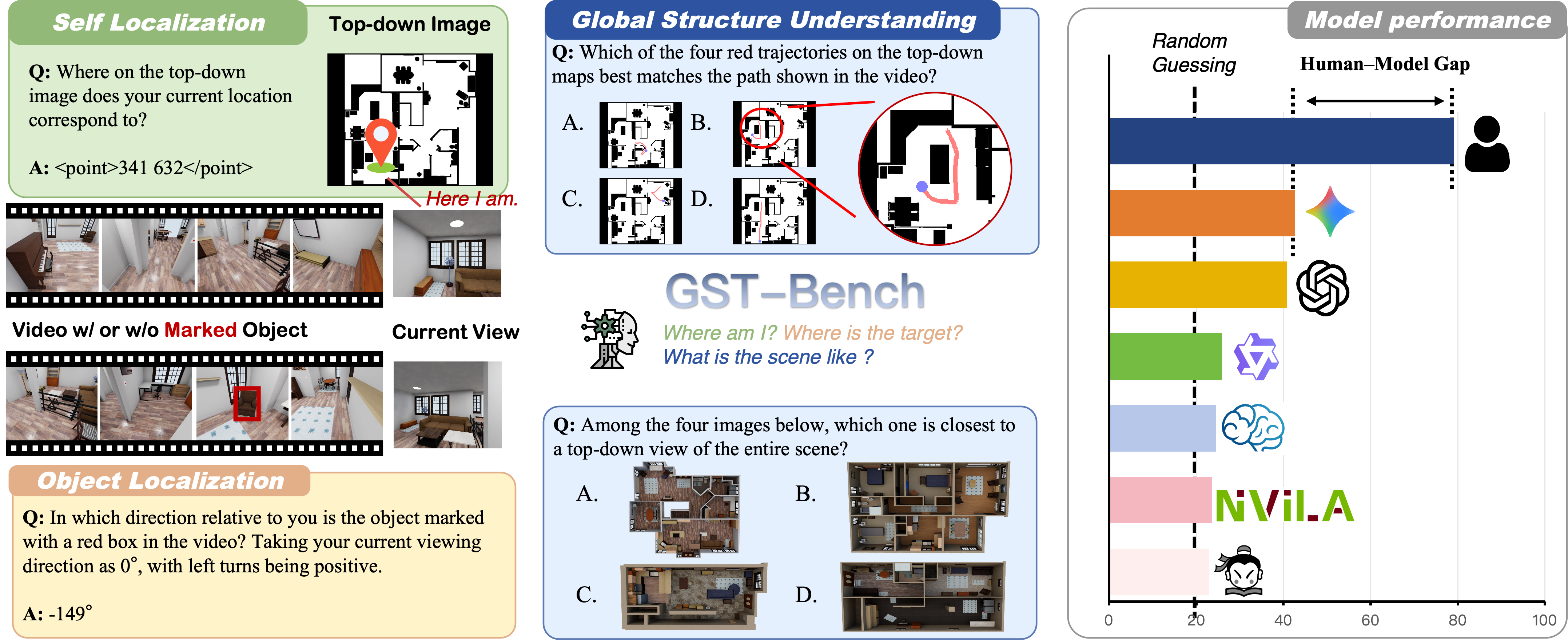} % Reduce the figure size so that it is slightly narrower than the column.
\caption{Overview of \textbf{GST-Bench}. We evaluate global spatial awareness along three core competencies — self localization (Where am I?), object localization (Where is the target?), and scene structure understanding (What does the scene look like?). Given an egocentric exploration video and a novel query view, models must localize themselves, reason about targets unseen in the current view and map observations onto a global top-down representation. As shown on the right, even the strongest VLM (Gemini-3-Pro, 42.68) falls far below the human baseline (79.08).}
\label{fig2}
\end{figure*}

Vision-Language Models (VLMs)~\cite{Gemini2.5,qwen3-vl,gpt5,Seed1.8} have achieved rapid progress in visual perception and reasoning. However, to empower VLMs to effectively interact with the physical world, \textbf{spatial intelligence} is indispensable — the ability to perceive surrounding layouts, track egocentric position, and maintain a globally consistent spatial understanding over time. Consider a household robot tasked with retrieving an item: it must incrementally build a scene representation from egocentric exploration video, remember where objects were encountered, and infer their relative locations regardless of whether they remain in the current field of view. Such spatial reasoning, which humans perform effortlessly, remains a critical challenge for current VLMs.

However, most existing spatial benchmarks are confined to \textbf{single-image}~\cite{whatsup,vsr,cambrian}
or \textbf{multi-image}~\cite{mmsi,spar} settings with typically 2–3 views, focusing on local 
spatial properties (e.g., object size, relative depth) rather than global scene 
awareness that requires coherent spatio-temporal understanding. While recent \textbf{video-based} benchmarks~\cite{mmsi-video,VSI-bench} begin to explore global spatial understanding, several gaps remain: (1) they do not disentangle single-frame solvable tasks from those requiring cross-frame global reasoning; (2) spatial metrics remain coarse-grained, such as front/back/left/right, rather than precise angles; and (3) they lack explicit global scene representations such as top-down images to evaluate global-local correspondence between the scene and individual frames.

To address these limitations, we design GST-Bench around four principles: 
\textbf{First}, the target object is guaranteed to be invisible from the query viewpoint, and questions answerable from a single video frame (e.g., object size) are explicitly excluded, ensuring that every task requires integrating information across views. 
\textbf{Second}, rather than relying on coarse-grained judgments such as ``which of A and B is closer'' or ``is A on the left or the right,'' GST-Bench computes its metrics over precise numerical answers for quantities such as distances and angles, enabling fine-grained and unambiguous evaluation.
\textbf{Third}, we introduce top-down images to probe a model's ability to map a long, cross-view video stream onto a global scene representation. \textbf{Finally}, every image representing the embodied agent's query viewpoint is sampled from an independent point off the video's camera trajectory, further increasing the difficulty.

Our task taxonomy is driven by three fundamental questions an embodied agent must resolve when reasoning about a scene globally — Where am I? (\textbf{self localization}), Where is the target? (\textbf{object localization}), and What does the scene look like? (\textbf{scene structure understanding}). We organize these three dimensions into 12 subtasks, yielding 2,762 human-verified questions with 6,790 minutes of cumulative QA-level video input.

To construct GST-Bench, we introduce an automatic pipeline built upon OmniGibson simulation with BEHAVIOR-1K~\cite{behavior}. The pipeline operates across 50 diverse indoor scenes, generating varied camera trajectories per scene alongside out-of-trajectory novel viewpoints. It systematically records the 3D coordinates of objects and viewpoints, projects them onto global top-down images, and automatically constructs QA pairs — while strictly filtering out any samples in which the target object is visible from the query viewpoint, ensuring that correct answers cannot be inferred from a single frame alone.

We evaluate 22 state-of-the-art VLMs, spanning open-source models, proprietary 
models, and embodied-understanding models such as RoboBrain~\cite{RoboBrain2.5} and Robix~\cite{robix}. The strongest 
zero-shot model, Gemini-3-Pro~\cite{google2025gemini3}, achieves only a score of 42.68, far below the human baseline of 
79.08. Open-source models perform substantially worse: even the strongest ones, InternVL3.5-38B~\cite{Internvl3.5}, remain around 30 points, while many models score close to random guessing. These results reveal that current VLMs, regardless of model family or training recipe, still exhibit severe limitations in global spatial awareness.

Furthermore, to disentangle global reasoning ability from local perception, we 
construct controlled local variants by making the target object visible in the 
query viewpoint, so that models need only reason about a single image rather than integrate information across the full video. Results reveal two qualitatively different bottlenecks: proprietary models such as Gemini-3-Pro exhibit gains of up to 39 points when the task is reduced to local perception, indicating that their failure is one of cross-frame spatial reasoning rather than single-image understanding; open-source models, by contrast, fail to improve consistently 
even in the local setting, exposing deficiencies at both the perception and 
reasoning stages.

Finally, leveraging the same data construction pipeline, we collect GST-Train, a large-scale training dataset built upon BEHAVIOR-1K, ArtVIP, HyperSim, and 
several internal simulation scenes, tailored for global spatial reasoning. 
Fine-tuning Qwen3-VL-8B on GST-Train yields a 27.63-point improvement on 
GST-Bench, raising the score from 25.89 to 53.52 and surpassing all 
proprietary models under zero-shot evaluation, demonstrating that the 
global spatial reasoning gap can be effectively narrowed through targeted 
training data.

% summarize the contributions
In summary, our contributions are as follows:
\begin{itemize}
    \item We introduce \textbf{GST-Bench}, a video-based VQA benchmark for evaluating global spatial awareness in VLMs. GST-Bench requires models to integrate long-horizon egocentric observations, reason from novel viewpoints, and map video evidence onto explicit top-down scene representations.
    \item We design a taxonomy of 12 subtasks covering three core competencies of embodied spatial intelligence: self localization, object localization, and scene structure understanding. All object-localization tasks enforce that the target object is invisible from the current view, ensuring that correct answers require cross-frame global reasoning rather than single-frame perception.
    \item We conduct a comprehensive evaluation of 22 state-of-the-art VLMs and reveal a substantial gap between current models and humans. Gemini-3-Pro, the strongest zero-shot model, achieves only 42.68 compared with the human score of 79.08, while most open-source models remain around 20--30 points and often close to random guessing.
    \item We further construct controlled local variants and \textbf{GST-Train} to diagnose and mitigate this limitation. Our analysis shows that proprietary models mainly fail at cross-frame spatial integration, whereas open-source models struggle with both local perception and global integration. Fine-tuning Qwen3-VL-8B on GST-Train improves its score from 25.89 to 53.52, demonstrating the effectiveness of targeted supervision for global spatial reasoning.
\end{itemize}

\section{Related Work}
\label{sec:Related Work}

\paragraph{Spatial Understanding from Single or Few Views.}
Early benchmarks formulate spatial understanding as categorical judgment, asking
models to verify preposition- or relation-level statements such as
\textbf{left of} and \textbf{under}~\cite{whatsup,vsr}. CV-Bench broadens this
evaluation to relation and counting queries for 2D understanding, together with
depth ordering and relative distance for 3D understanding~\cite{cambrian}.
Subsequent work moves beyond these basic task categories toward more 3D-aware
reasoning. 3DSRBench evaluates height, location, orientation, and multi-object
3D relations, including robustness to uncommon camera viewpoints~\cite{3dsrbench},
and Spatial457 further scales diagnostic difficulty to 6D spatial relationships
and collision prediction~\cite{spatial457}.

To go beyond the ambiguity of a single viewpoint, recent work introduces
additional views or multi-image inputs. MM-Spatial and SPAR support single- and
multi-view inputs for tasks such as spatial relationship prediction, metric
estimation, and 3D grounding~\cite{mmspatial,spar}, while BLINK exposes
complementary perceptual gaps through classic vision tasks including relative
depth and multi-view reasoning~\cite{blink}. Another line specifically targets
cross-view competence: ViewSpatial-Bench evaluates reasoning across camera- and
human-centric frames of reference~\cite{viewspatial}, and MMSI-Bench requires
models to integrate multiple images, track camera or object motion, and relate
entities that may never co-occur in a single frame~\cite{mmsi}. Despite this progress, image and multi-image settings still primarily test
localized spatial understanding from a small number of discrete views, often
covering only partial regions of a scene, making it difficult to evaluate whether a model can form and
maintain a coherent global representation of the entire environment.

\paragraph{Spatial Understanding from Video.}
Video-based benchmarks move spatial reasoning toward broader scene-level
understanding, where models must integrate observations over time to reason
about room-scale or even cross-room spatial structure. Built on egocentric
indoor videos, VSI-Bench evaluates configurational reasoning, metric estimation,
and spatiotemporal understanding over 3D spaces~\cite{VSI-bench}. VSI-Super
further stresses long-horizon processing through visual spatial recall and
continual counting over arbitrarily long streams~\cite{vsisuper}. Complementarily, MMSI-Video-Bench provides a human-annotated benchmark for
video-based spatial intelligence, spanning perception, planning, prediction, and
cross-video reasoning~\cite{mmsi-video}. Embodied QA
benchmarks such as OpenEQA evaluate episodic memory and active exploration in
real-world environments~\cite{openeqa}, while OST-Bench studies online
spatio-temporal understanding from the perspective of an actively exploring
agent~\cite{ostbench}. More recent work such as STI-Bench explicitly examines
precise spatial-temporal quantities, including pose, displacement, speed,
acceleration, and trajectory understanding~\cite{stibench}.

Although these benchmarks begin to probe global spatial understanding, three
gaps remain. \textbf{First}, many evaluations do not explicitly separate
questions that can be answered from a single informative frame from those that
strictly require cross-frame integration. For example, tasks such as object-size estimation in VSI-Bench or existence
queries in OST-Bench may be answerable from a single frame in the video. While this broadens benchmark coverage, it weakens
the targeted evaluation of cross-frame global spatial reasoning. \textbf{Second},
while recent benchmarks increasingly include numerical quantities, direction-related
spatial reasoning is still often evaluated through coarse categorical relations
such as left/right or front/back, rather than precise angular measurements.
\textbf{Third}, existing benchmarks rarely evaluate the correspondence between
egocentric video observations and an explicit global visual representation, such
as a top-down image of the scene. GST-Bench is designed to close these gaps:
each task is constructed to require global cross-frame understanding rather than
being answerable from any single input frame, answers are scored against precise
numerical quantities, and top-down images are introduced to directly assess
whether egocentric observations can be organized into a globally consistent
scene representation.

\section{GST-Bench}
\label{sec:GST-Bench}

\begin{figure*}[!t]
\centering
\includegraphics[width=\textwidth]{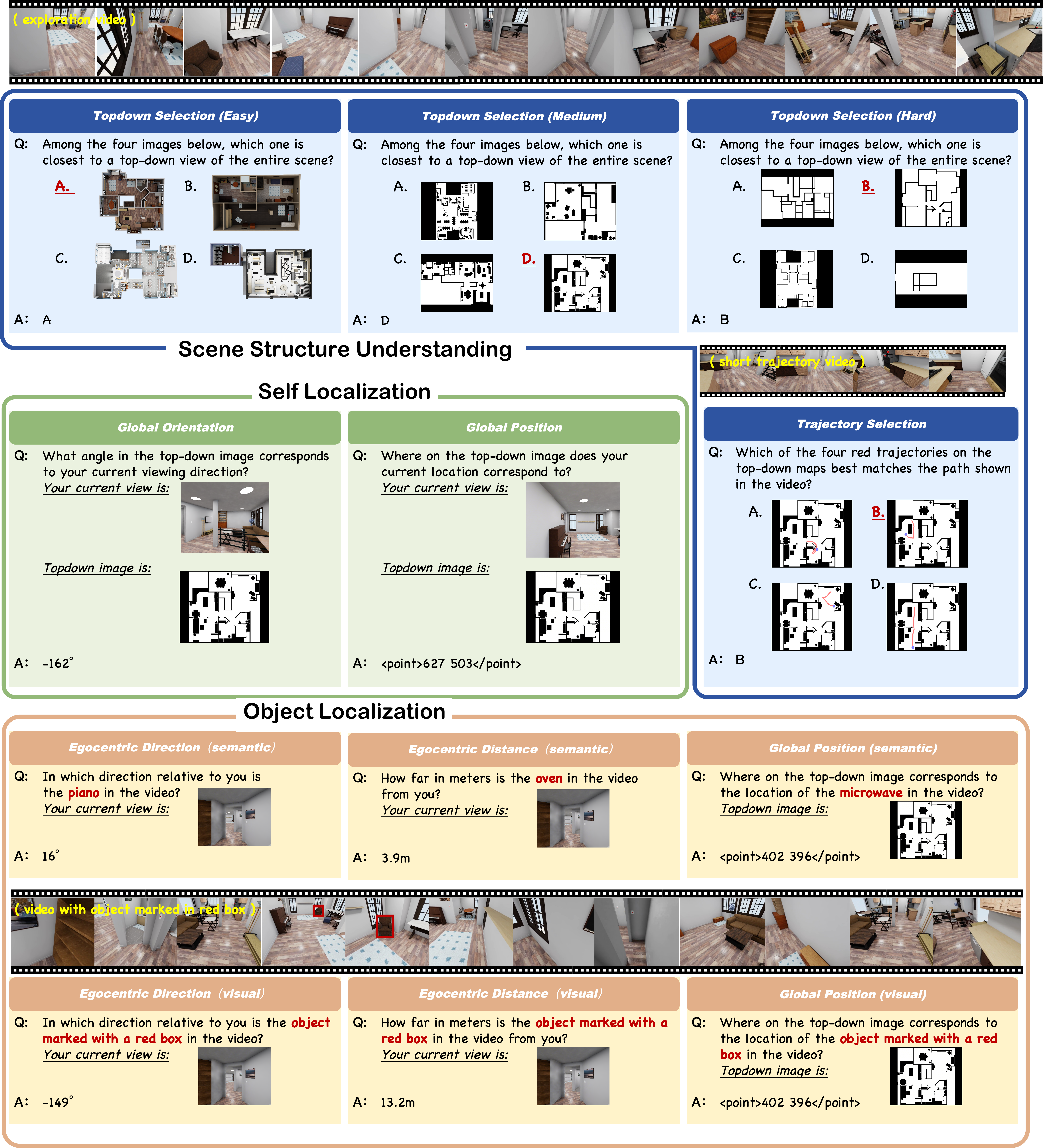} 
    \caption{Representative GST-Bench samples from each of the twelve task types, organized by the three core competencies: self localization, object localization, and scene structure understanding.}
    \label{fig:all_task}
\end{figure*}

GST-Bench evaluates whether VLMs can build a globally consistent spatial representation from long-horizon egocentric video. Unlike spatial benchmarks centered on single images or a small number of views, GST-Bench requires models to integrate observations over time, reason from novel viewpoints that do not appear in the input video, and align egocentric evidence with explicit top-down scene representations. We organize the benchmark around three core competencies required by embodied agents: \textbf{self localization}, \textbf{object localization}, and \textbf{scene structure understanding}. These competencies are instantiated as twelve task types, as shown in Figure~\ref{fig:all_task}.

\subsection{Task Inputs}
\label{sec:task_inputs}

GST-Bench includes the following five types of visual inputs.

\paragraph{Exploration Video.}
An exploration video provides a long-horizon egocentric traversal of the entire scene. It serves as the primary source from which the model must accumulate spatial memory, infer object locations, and build a global understanding of the environment.

\paragraph{Object-Annotated Video.}
An object-annotated video augments the exploration video with a red bounding box overlaid on every frame in which a designated target instance is visible. This input is used in the visual modality of object-localization tasks, where the target is specified visually rather than by category name.

\paragraph{Short Trajectory Video.}
A short trajectory video is a clip segmented from an independently collected exploration trajectory. It is used in the trajectory selection task, where the model must match the observed egocentric motion to one of several candidate trajectories on a top-down image.

\paragraph{Top-Down Images.}
Top-down images provide explicit global scene representations. We render three abstraction levels. The \textit{easy} version is a photo-realistic bird's-eye-view rendering with the ceiling removed, preserving object textures, colors, and spatial arrangements. The \textit{medium} version is an occupancy-style map that retains the spatial footprints of objects and walls while removing appearance details. The \textit{hard} version is a bare floor plan that retains only wall structures and removes all objects. These variants allow us to control how much global visual information is available.

\paragraph{Current View.}
The current view is a novel egocentric image rendered from a viewpoint that is absent from the input video trajectory. It is used as the query viewpoint for self-localization and object-localization tasks. For all object-localization tasks, the target object is explicitly required to be invisible from the current view, ensuring that the task cannot be solved from local perception alone.

\subsection{Task Taxonomy}
\label{sec:task_taxonomy}

GST-Bench instantiates the above inputs into twelve task types organized by three core competencies.

\paragraph{Self Localization.}
Self localization evaluates whether a model can align a novel egocentric view with a global top-down representation of the scene.
\begin{itemize}
    \item \textbf{Global Orientation.} \textit{Input: exploration video, medium-level top-down image, and novel current view.} Infer the camera orientation of the current view on the top-down image.
    \item \textbf{Global Position.} \textit{Input: exploration video, medium-level top-down image, and novel current view.} Infer the camera position of the current view on the top-down image.
\end{itemize}

\paragraph{Object Localization.}
Object localization evaluates whether a model can infer the location of a target object that was observed during exploration but is absent from the current view. We consider two target-specification modalities: in the \textbf{semantic} modality, the target is specified by its category name; in the \textbf{visual} modality, the target is specified by bounding-box annotations in an object-annotated video. For all object-localization tasks, the target object is absent from the current view, ensuring that the task requires cross-frame spatial reasoning.
\begin{itemize}
    \item \textbf{Egocentric Direction (semantic).} \textit{Input: exploration video, target category name, and novel current view.} Infer the egocentric direction from the current viewpoint to the target object.
    \item \textbf{Egocentric Direction (visual).} \textit{Input: object-annotated video and novel current view.} Infer the egocentric direction from the current viewpoint to the visually specified target object.
    \item \textbf{Egocentric Distance (semantic).} \textit{Input: exploration video, target category name, and novel current view.} Estimate the metric distance from the current viewpoint to the target object.
    \item \textbf{Egocentric Distance (visual).} \textit{Input: object-annotated video and novel current view.} Estimate the metric distance from the current viewpoint to the visually specified target object.
    \item \textbf{Global Position (semantic).} \textit{Input: exploration video, target category name, medium-level top-down image, and novel current view.} Predict the target object's position on the top-down image.
    \item \textbf{Global Position (visual).} \textit{Input: object-annotated video, medium-level top-down image, and novel current view.} Predict the visually specified target object's position on the top-down image.
\end{itemize}

\paragraph{Scene Structure Understanding.}
Scene structure understanding evaluates whether a model can organize egocentric observations into a global representation of the environment.
\begin{itemize}
    \item \textbf{Top-Down Selection (easy).} \textit{Input: exploration video and four easy-level candidate top-down images.} Select the top-down image that matches the explored scene.
    \item \textbf{Top-Down Selection (medium).} \textit{Input: exploration video and four medium-level candidate top-down images.} Select the top-down image that matches the explored scene.
    \item \textbf{Top-Down Selection (hard).} \textit{Input: exploration video and four hard-level candidate top-down images.} Select the top-down image that matches the explored scene.
    \item \textbf{Trajectory Selection.} \textit{Input: short trajectory video and a medium-level top-down image overlaid with four candidate trajectories.} Select the trajectory that matches the observed egocentric motion.
\end{itemize}

\subsection{Data Generation Pipeline}
\label{sec:pipeline}

\begin{figure*}[htbp]
\centering
\includegraphics[width=\textwidth]{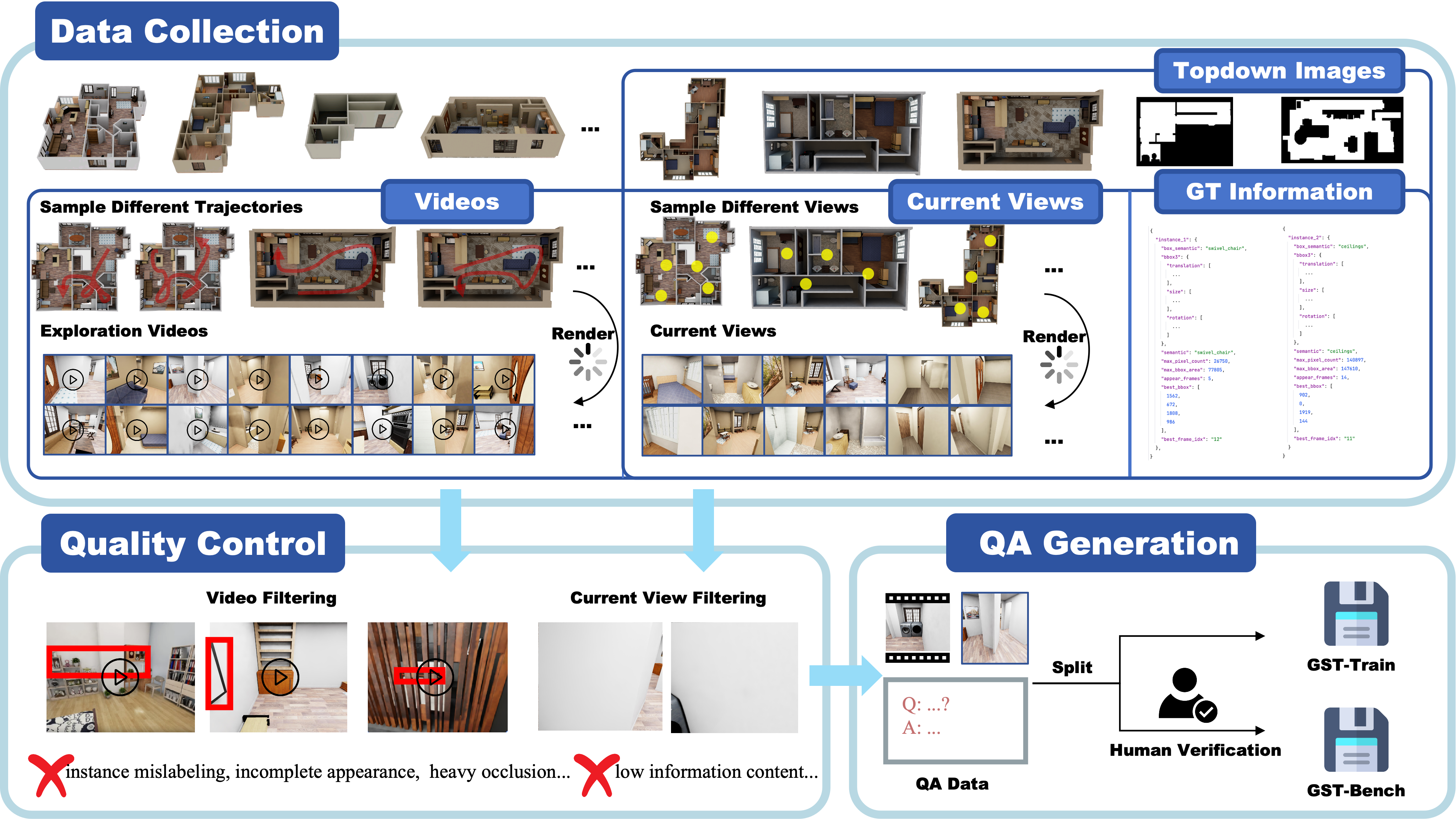} 
    \caption{Overview of the automatic data generation pipeline. Starting from diverse simulation scenes, the pipeline produces exploration videos, object-annotated videos, novel-viewpoint images, top-down images at three abstraction levels, and template-based QA pairs, followed by automated filtering and human verification.}
    \label{fig:pipeline}
\end{figure*}

Figure~\ref{fig:pipeline} summarizes our automatic data generation pipeline. The pipeline is designed with three characteristics. 
\textbf{First, it enforces global reasoning by construction.} Object-localization questions require the target object to be visible in the exploration video but invisible from the current view, and the current view is sampled outside the video trajectory. Therefore, models cannot solve the task by recognizing the target in a single image or by directly matching the query image to a video frame.
\textbf{Second, it leverages the scalability and controllability of simulation.} In simulation, we can sample diverse video trajectories and arbitrary numbers of viewpoints within each scene, generate query viewpoints independent of the exploration trajectory, and flexibly vary camera poses, orientations, and other rendering conditions. At the same time, camera poses, object poses, object visibility, distances, angles, and top-down projections are all available from scene geometry, enabling precise and scalable answer generation.
\textbf{Third, it combines automatic generation with human verification for benchmark quality.} For GST-Bench, annotators verify the answerability of each sample from two aspects: whether the target object is identifiable in the video, and whether the off-trajectory current view can still be localized from the exploration video. These checks ensure that the benchmark is both challenging and reliably answerable.

\paragraph{Scene Preparation.}
We build GST-Bench and GST-Train from diverse indoor simulation assets, including BEHAVIOR-1K~\cite{behavior}, HyperSim~\cite{hypersim}, ArtVIP~\cite{artvip}, and additional simulation scenes. The key goal is to cover varied room layouts, object arrangements, and navigable structures, so that global spatial reasoning is evaluated across full-scene environments rather than repeated local configurations. The evaluation scenes used for GST-Bench are held out from GST-Train to prevent scene-level leakage.

\paragraph{Exploration Video Generation.}
Exploration videos are designed to provide broad scene coverage for long-horizon spatial memory. For each scene, we sample spatially distributed viewpoints over the navigable area, connect them into an efficient traversal trajectory, and render egocentric RGB videos along the path. To avoid relying on a single canonical route, we generate multiple trajectories with perturbed viewpoints and different starting locations. The resulting videos serve as the primary input from which models must build a global representation of the scene.

\paragraph{Object-Annotated Video Generation.}
Object-annotated videos support the visual modality of object localization, where the target is specified by appearance rather than by category name. Using instance and semantic segmentation rendered from simulation, we identify visible object instances and verify their visibility and category correctness. For each selected target instance, we overlay a red bounding box on all frames where it appears. This enables instance-level target specification while still requiring the model to remember the target's global location after it disappears from the current view.

\paragraph{Short Trajectory Video Generation.}
Short trajectory videos are generated to evaluate whether models can align egocentric motion with a global map. We collect additional trajectories independently from the main exploration videos and segment them into short continuous clips. These clips provide local motion observations, while the answer requires selecting the corresponding trajectory among candidates overlaid on a top-down image.

\paragraph{Top-Down Image Generation.}
Top-down images provide explicit global scene representations and allow us to evaluate egocentric-to-global alignment. For each scene, we render three abstraction levels: an \textit{easy} photo-realistic bird's-eye-view image, a \textit{medium} occupancy-style map that preserves object and wall footprints, and a \textit{hard} floor-plan image that retains only wall structures. These variants support both scene-level matching and fine-grained localization on a global map.

\paragraph{Current View Generation.}
Current views are central to preventing shortcut solutions. Instead of sampling query images from the exploration video, we generate them from off-trajectory viewpoints over the navigable region with independently sampled camera orientations. A query view is retained only when it has sufficient visual overlap with the exploration video to make localization possible. For object-localization tasks, we additionally require the target object to be absent from this current view, forcing models to infer the target's location from prior video observations.

\paragraph{QA Generation and Automatic Filtering.}
We combine the generated inputs with task-specific templates to produce QA pairs for all twelve task types. Because the data are generated in simulation, answers are computed directly from geometry: egocentric directions and distances from camera-object poses, point locations from top-down projections, and multiple-choice labels from scene or trajectory identities. Automatic filtering removes samples with invalid projections, ambiguous object visibility, insufficient query-view overlap, degenerate choices, or violations of the target-invisible constraint.

\paragraph{Human Verification.}
For GST-Bench, every evaluation sample is further manually verified. Annotators focus on answerability rather than merely checking generated labels: they confirm that the target object can be recognized in the exploration or object-annotated video, and that the current view, although outside the video trajectory, contains enough overlap with the video for a human to localize it within the scene. Samples failing these checks are discarded, yielding 2{,}762 human-verified questions for the final benchmark.

\section{Evaluation on GST}
\label{sec:Evaluation on GST}

\subsection{Evaluation Setup}

\paragraph{Baseline models.}
We comprehensively evaluate 22 state-of-the-art VLMs, spanning proprietary models, open-source models, and embodied-understanding models, across diverse model families, parameter scales (from 2B to 38B), and training recipes.
For proprietary models, we consider Gemini-2.5-Pro~\cite{Gemini2.5}, Gemini-3-Pro~\cite{google2025gemini3}, GPT-5~\cite{gpt5}, GPT-4o~\cite{Gpt-4o}, and Seed1.8~\cite{Seed1.8}.
For open-source models, we include representatives from LLaVA-OneVision-1.5~\cite{LLaVA-OneVision-1.5} (4B, 8B), the Qwen3-VL family~\cite{qwen3-vl} (2B, 4B, 8B, 32B), the InternVL3.5 family~\cite{Internvl3.5} (2B, 4B, 8B, 38B), and NVILA~\cite{Nvila} (8B, 15B), covering a wide range of architectural designs and scaling behaviors.
For embodied-understanding models, we further incorporate Cosmos-Reason2~\cite{cosmosreason2} (2B, 8B), RoboBrain2.5-8B~\cite{RoboBrain2.5}, and Robix~\cite{robix} (7B, 32B), which are specifically tailored for grounded spatial reasoning and action understanding in physical environments.
Additionally, we include Qwen3-VL-8B fine-tuned on our proposed GST-Train dataset (Ours), to validate the effectiveness of our approach.
All models are evaluated under zero-shot settings using their officially recommended prompt templates. To ensure fair comparison and reproducibility, we adopt greedy decoding throughout and report results without any task-specific fine-tuning.

\paragraph{Metric Design.}
GST-Bench comprises four distinct question formats: distance prediction, angle prediction, point prediction, and multiple-choice questions. We adopt tailored evaluation metrics for each format.

For \textbf{distance prediction.}, we employ Mean Relative Accuracy ($\mathcal{MRA}$) following the protocol of VSI-Bench~\cite{VSI-bench}. Given a model's prediction $\hat{y}$, ground truth $y$, and a set of confidence thresholds $\mathcal{C}_d = \{0.50, 0.55, \ldots, 0.95\}$, $\mathcal{MRA}$ averages the indicator of whether the relative error falls below $1 - \theta$ across all thresholds:
\begin{equation}
    \mathcal{MRA} = \frac{1}{|\mathcal{C}_d|} \sum_{\theta \in \mathcal{C}_d} \mathbbm{1}\left(\frac{|\hat{y} - y|}{y} < 1 - \theta\right).
    \label{eq:mra}
\end{equation}
For \textbf{angle prediction}, we compute the angular error as the circular distance between the predicted angle $\hat{\alpha}$ and the ground-truth angle $\alpha$:
\begin{equation}
e_a = \min\left(|\hat{\alpha}-\alpha|,\; 360^{\circ} - |\hat{\alpha}-\alpha|\right),
\end{equation}
where all angles are expressed in degrees and normalized to a common range. The metric is defined as the mean accuracy over a set of tolerance thresholds $\mathcal{C}_a = \{15^{\circ}, 30^{\circ}, 45^{\circ}\}$:

\begin{equation}
    \mathrm{Acc}_{\text{angle}} = \frac{1}{|\mathcal{C}_a|} \sum_{\tau \in \mathcal{C}_a} \mathbbm{1}\left(e_a < \tau\right).
    \label{eq:acc_angle}
\end{equation}
For \textbf{point prediction}, we compute the Euclidean distance $e_p = \|\hat{\mathbf{p}} - \mathbf{p}\|_2$ between the predicted point $\hat{\mathbf{p}}$ and the ground-truth point $\mathbf{p}$. The metric is defined as the mean accuracy over a set of distance thresholds $\mathcal{C}_p = \{100, 150, 200, 250, 300\}$ (in pixels):
\begin{equation}
    \mathrm{Acc}_{\text{point}} = \frac{1}{|\mathcal{C}_p|} \sum_{\tau \in \mathcal{C}_p} \mathbbm{1}\left(e_p < \tau\right).
    \label{eq:acc_point}
\end{equation}
For \textbf{multiple-choice questions}, we directly adopt standard accuracy as the evaluation metric.

The overall GST-Bench score is defined as the arithmetic mean of the individual scores across all 12 subtasks:
\begin{equation}
    \mathrm{Score} = \frac{1}{12} \sum_{i=1}^{12} s_i,
    \label{eq:overall}
\end{equation}
where $s_i$ denotes the evaluation score of the $i$-th subtask under its corresponding metric.

\begin{table*}[!t]
\centering
\caption{Performance of various models on GST-Bench across three core competencies: \textbf{Object Localization}, \textbf{Self Localization}, and \textbf{Scene Structure Understanding}. The \textbf{Avg.} is the arithmetic mean of all 12 subtask scores. The highest, second-highest, and third-highest scores in each column are marked in \colorbox{red!28}{light red}, \colorbox{orange!28}{light orange}, and \colorbox{yellow!35}{light yellow}, respectively, excluding Qwen3-VL-8B (fine-tuned), which is fine-tuned on our proposed training set GST-Train. ED = Ego. Direction, EDist = Ego. Distance, GP = Global Position, Pos = Top-Down Position, Ori = Top-Down Orientation, TDS = Top-Down Selection, Traj = Trajectory Selection. $_v$ / $_s$ = visual / semantic modality.}
\label{tab:main_results}
\setlength{\tabcolsep}{2.0pt}
\renewcommand{\arraystretch}{1.05}
\scriptsize
\begin{tabular}{@{}l *{6}{c} *{2}{c} *{4}{c} c@{}}
\toprule
& \multicolumn{6}{c}{Object Localization} & \multicolumn{2}{c}{Self Localization} & \multicolumn{4}{c}{Scene Structure Understanding} & \\
\cmidrule(lr){2-7} \cmidrule(lr){8-9} \cmidrule(lr){10-13}
Method
& ED$_v$ & EDist$_v$ & GP$_v$ & ED$_s$ & EDist$_s$ & GP$_s$
& Pos & Ori
& TDS$_E$ & TDS$_M$ & TDS$_H$ & Traj & Avg. \\
& \footnotesize mAcc & \footnotesize MRA & \footnotesize mAcc & \footnotesize mAcc & \footnotesize MRA & \footnotesize mAcc
& \footnotesize mAcc & \footnotesize mAcc
& \footnotesize Acc & \footnotesize Acc & \footnotesize Acc & \footnotesize Acc & \\
\midrule
\multicolumn{14}{@{}l}{\cellcolor{gray!12}\textit{Proprietary Models}} \\
Gemini-2.5-Pro     & 18.60 & \rka{42.00} & \rkc{31.80} & \rkb{23.08} & \rka{35.40} & 32.40 & \rkb{30.47} & \rka{21.52} & \rkc{97.69} & \rkb{80.00} & 41.01 & \rkc{37.44} & \rkb{40.95} \\
Gemini-3-Pro       & \rkb{21.73} & \rkc{25.35} & \rka{42.23} & \rkc{22.11} & \rkb{27.39} & \rka{43.67} & \rka{37.47} & 16.89 & \rkb{98.15} & \rka{82.73} & \rka{52.25} & \rka{42.18} & \rka{42.68} \\
GPT-5              & \rka{27.98} & \rkb{27.35} & \rkb{34.28} & \rka{31.09} & \rkc{27.39} & \rkb{43.23} & 22.33 & 16.74 & \rkc{97.69} & \rkc{79.55} & \rkc{43.26} & \rkb{39.34} & \rkc{40.85} \\
GPT-4o             & 19.64 & 10.09 & 22.36 & 13.46 & 10.13 & 25.24 & 24.03 & 15.99 & 81.02 & 71.36 & \rkb{45.51} & 25.12 & 30.33 \\
Seed1.8            & 16.37 & 12.43 & 27.96 & 20.35 & 13.32 & \rkc{36.07} & \rkc{28.45} & 17.79 & \rka{99.07} & 67.27 & 37.64 & 31.75 & 34.04 \\
\midrule
\multicolumn{14}{@{}l}{\cellcolor{gray!12}\textit{Open-Source Models}} \\
LLaVA-OV-1.5-4B    & 17.41 & 4.17  & 14.28 & 17.63 & 4.34  & 18.86 & 20.23 & 16.29 & 49.54 & 37.27 & 27.53 & 25.59 & 21.10 \\
LLaVA-OV-1.5-8B    & 17.71 & 5.52  & 14.22 & 18.11 & 3.94  & 10.83 & 13.33 & 16.89 & 52.78 & 33.18 & 21.91 & 26.54 & 19.58 \\
Qwen3-VL-2B        & 17.41 & 4.13  & 9.73  & 16.19 & 6.55  & 18.08 & 20.93 & \rkc{18.39} & 88.89 & 25.91 & 21.35 & 24.17 & 22.64 \\
Qwen3-VL-4B        & 15.33 & 4.09  & 17.46 & 16.66 & 6.55  & 12.66 & 19.61 & 18.09 & 94.44 & 34.09 & 22.47 & 26.07 & 23.96 \\
Qwen3-VL-8B        & 18.75 & 5.57  & 17.35 & 12.50 & 11.59 & 13.01 & 24.81 & 17.64 & 96.76 & 42.27 & 25.28 & 25.12 & 25.89 \\
Qwen3-VL-32B       & 16.67 & 13.96 & 18.05 & 15.22 & 16.19 & 26.55 & 24.03 & 17.64 & 94.44 & 60.45 & 32.58 & 29.38 & 30.43 \\
InternVL3.5-2B     & 18.45 & 2.70  & 9.50  & 18.11 & 4.56  & 13.97 & 12.25 & 17.34 & 59.72 & 29.09 & 25.84 & 25.12 & 19.72 \\
InternVL3.5-4B     & 15.63 & 4.96  & 8.91  & 12.02 & 5.40  & 5.15  & 20.62 & 16.29 & 66.67 & 30.91 & 25.28 & 26.07 & 19.83 \\
InternVL3.5-8B     & 17.71 & 5.57  & 19.06 & 16.67 & 8.36  & 14.85 & 18.84 & 13.90 & 75.00 & 35.00 & 25.84 & 25.12 & 22.99 \\
InternVL3.5-38B    & \rkc{21.28} & 11.13 & 20.29 & 20.03 & 10.44 & 33.10 & 22.09 & \rkb{20.63} & 75.00 & 68.64 & 36.52 & 29.38 & 30.71 \\
NVILA-8B           & 18.15 & 7.22  & 15.87 & 16.02 & 7.52  & 26.64 & 6.67  & 16.44 & 79.17 & 34.55 & 29.78 & 25.59 & 23.64 \\
NVILA-15B          & 15.47 & 4.74  & 8.85  & 16.66 & 3.10  & 10.13 & 8.06  & 17.79 & 74.07 & 40.00 & 32.02 & 22.27 & 21.10 \\
\midrule
\multicolumn{14}{@{}l}{\cellcolor{gray!12}\textit{Embodied-Understanding Models}} \\
Cosmos-Reason2-2B  & 15.18 & 3.91  & 8.38  & 14.10 & 7.65  & 8.47  & 17.75 & 16.74 & 78.70 & 24.09 & 26.40 & 18.48 & 19.99 \\
Cosmos-Reason2-8B  & 13.39 & 6.96  & 10.86 & 16.83 & 7.61  & 10.57 & 19.77 & 14.50 & 84.26 & 28.64 & 23.60 & 22.75 & 21.64 \\
RoboBrain2.5-8B    & 20.24 & 10.39 & 14.04 & 19.23 & 11.28 & 14.06 & 20.31 & 16.74 & 80.56 & 37.27 & 26.97 & 24.17 & 24.61 \\
Robix-7B           & 17.41 & 12.35 & 24.13 & 18.43 & 12.52 & 21.83 & 22.95 & 15.99 & 88.43 & 53.18 & 35.96 & 25.59 & 29.06 \\
Robix-32B          & 18.90 & 24.87 & 21.95 & 18.91 & 20.18 & 21.05 & 22.48 & 17.94 & 84.72 & 43.64 & 30.90 & 25.59 & 29.26 \\
\midrule
\multicolumn{14}{@{}l}{\cellcolor{gray!12}\textit{Baseline}} \\
Random Guessing    & 14.88 & 23.22 & 15.04 & 17.31 & 20.40 & 14.15 & 16.51 & 15.55 & 23.15 & 27.73 & 24.72 & 27.49 & 20.01 \\
Human Level (20 samples/task) & 75.00 & 41.50 & 93.00 & 70.00 & 41.50 & 89.00 & 84.00 & 85.00 & 100.00 & 95.00 & 85.00 & 90.00 & 79.08 \\
\midrule
\multicolumn{14}{@{}l}{\cellcolor{gray!12}\textit{Ours (Fine-tuned on GST-Train)}} \\
Qwen3-VL-8B (fine-tuned) & 53.72 & 39.70 & 48.97 & 49.36 & 31.95 & 57.64 & 40.55 & 44.39 & 99.07 & 88.18 & 37.08 & 51.66 & 53.52 \\
\bottomrule
\end{tabular}
\end{table*}

\subsection{Main Results}
\label{sec:main_results}

Table~\ref{tab:main_results} reports the performance of all 22 evaluated VLMs on GST-Bench, together with the random-guessing and human-level baselines. Overall, GST-Bench proves highly challenging for current VLMs: even the strongest proprietary model trails human performance by over 36 points, the majority of open-source models perform near random, and embodied-tuned models show no advantage over their general-purpose counterparts. We elaborate on these findings below.

\paragraph{A huge gap between VLMs and humans.}
Human evaluators achieve an average score of 79.08, whereas the best model, Gemini-3-Pro, reaches only 42.68, leaving a gap of 36.4 points. The gap is pervasive across all three competencies, and is most pronounced on orientation estimation (Ori: 21.52 vs.\ 85.00) and global position estimation (GP$_v$: 42.23 vs.\ 93.00). One notable outlier is egocentric distance estimation, where the human baseline itself is low (\textasciitilde41 MRA) and the strongest model nearly matches it (EDist$_v$: 42.00 vs.\ 41.50). Rather than indicating model strength, this suggests that recovering absolute metric distance is intrinsically difficult for both humans and models.

\paragraph{Open-source models struggle on GST-Bench.}
While Qwen3-VL-32B (30.43) and InternVL3.5-38B (30.71) appear to lead the open-source tier, this advantage is mainly driven by the Top-Down Selection (easy/medium) subtasks; on the remaining ten tasks they remain within a few points of the random baseline. Four models — LLaVA-OV-1.5-8B (19.58), InternVL3.5-2B (19.72), InternVL3.5-4B (19.83), and Cosmos-Reason2-2B (19.99) — even fall \textbf{below} random, and nine of the seventeen open-source and embodied-understanding models score within three points of it. This indicates that, with the partial exception of coarse scene identification, current open-source VLMs lack any meaningful global spatial competence.

\paragraph{Proprietary models lead the benchmark, but only modestly.}
Gemini-3-Pro (42.68), Gemini-2.5-Pro (40.95), and GPT-5 (40.85) form a clear top tier, outperforming the best open-source model by roughly 10 points and roughly doubling the random-guessing baseline. Across all twelve subtasks, the per-task best is consistently held by a proprietary model. Nevertheless, even this top tier remains less than 55\% of human performance, leaving substantial headroom for the entire field.

\paragraph{Embodied-tuned models inherit, rather than mitigate, the global-reasoning blind spot.}
At the 8B scale, RoboBrain2.5-8B (24.61) and Cosmos-Reason2-8B (21.64) both fall \textbf{below} the general-purpose Qwen3-VL-8B (25.89); at the larger scale, Robix-32B (29.26) is on par with Qwen3-VL-32B (30.43). This is a striking negative result: despite being explicitly tuned for grounded spatial reasoning, embodied models inherit the same global-reasoning blind spot as their general-purpose backbones. We conjecture that current embodied post-training recipes emphasize local affordance and spatial relation prediction rather than the long-horizon spatial memory and cross-viewpoint alignment targeted by GST-Bench.

\begin{figure}[t]
\centering
\includegraphics[width=1\textwidth]{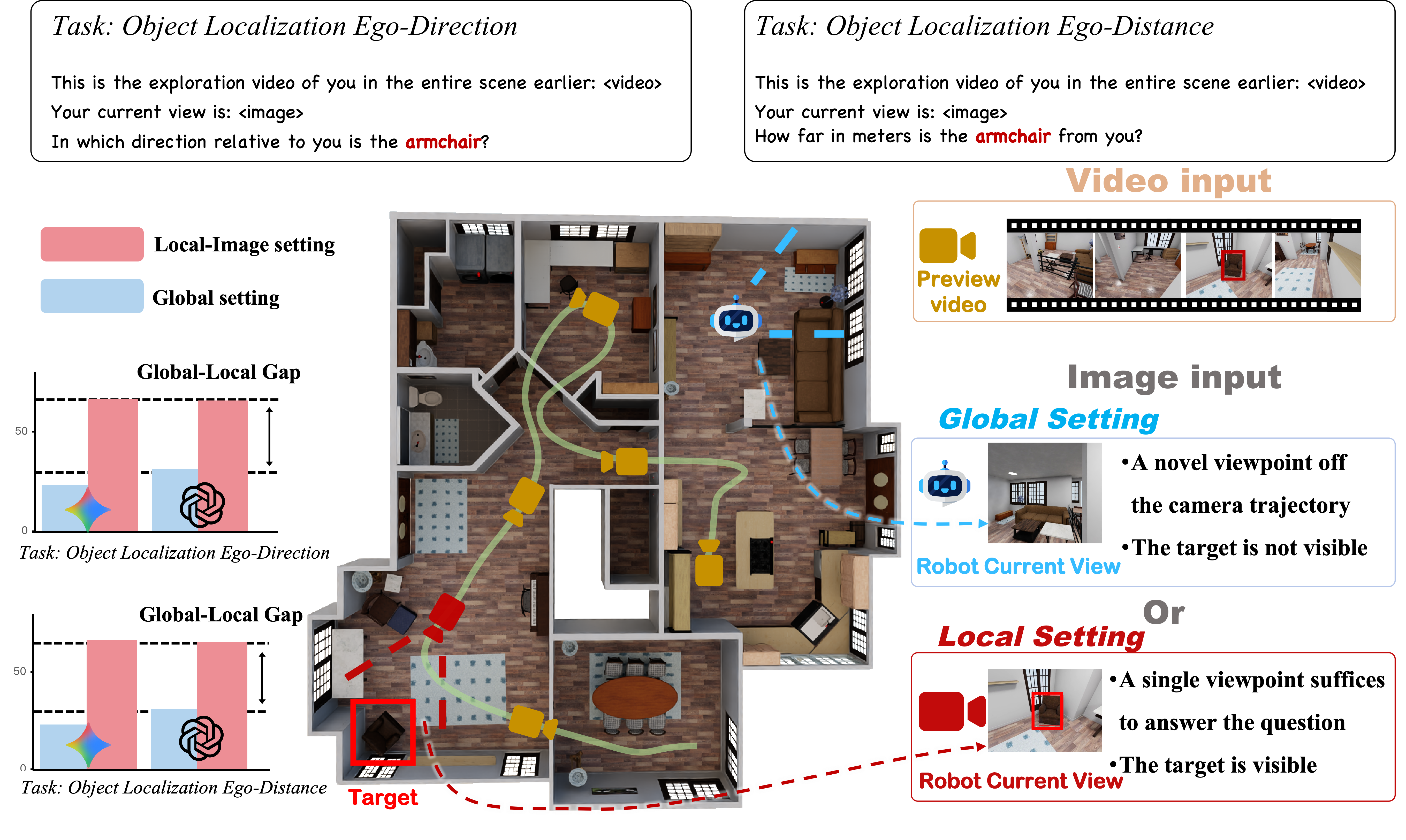}
\caption{Illustration of the three controlled settings used to disentangle global reasoning from local perception. Global (target absent from the current view, requires cross-frame integration), Local-Video (current view contains the target, exploration video redundant), and Local-Image (exploration video removed, reduced to single-image spatial understanding).}
\label{fig:ablation}
\end{figure}

\subsection{Disentangling Global Reasoning from Local Perception}
\label{sec:disentangle}

A natural concern is whether models fail on GST-Bench simply because they lack basic spatial perception, rather than because they cannot integrate information across the video. To disentangle these two factors, we construct controlled local variants that progressively remove the need for cross-frame reasoning while keeping the underlying spatial question identical. We focus on the Egocentric Direction (semantic) and Egocentric Distance (semantic) tasks, since semantic target specification eliminates visual-grounding confounds and isolates the spatial-reasoning component.

\begin{itemize}
    \item \textbf{Global.} The default GST-Bench setting: the target object is absent from the current view, and the model must reason over the exploration video to infer its spatial relation to the novel viewpoint.
    \item \textbf{Local-Video.} Identical to Global, except the current view is swapped for one in which the target object \textbf{is} visible. The exploration video is preserved but is now logically unnecessary — this isolates whether the model can ignore distractor video content when sufficient local evidence is available.
    \item \textbf{Local-Image.} The exploration video is further removed, leaving only the modified current view. The task reduces to single-image spatial perception.
\end{itemize}
Results are reported in Table~\ref{tab:disentangle_local}. Two qualitatively different bottlenecks emerge for proprietary and open-source models.

\begin{table}[!t]
\centering
\caption{Performance comparison under disentangled local settings for Egocentric Direction and Egocentric Distance. L-Video and L-Image denote Local-Video and Local-Image settings, respectively. Values in parentheses denote absolute gains over the Global setting.}
\label{tab:disentangle_local}
\setlength{\tabcolsep}{1.6pt}
\renewcommand{\arraystretch}{1.15}
\scriptsize

\begin{minipage}[t]{0.49\linewidth}
\centering
\textbf{(a) Egocentric Direction} \\[0.4em]
\begin{tabular}{@{}l ccc@{}}
\toprule
Method & Global & L-Video & L-Image \\
\midrule
\multicolumn{4}{@{}l}{\cellcolor{gray!12}\textit{Proprietary Models}} \\
Gemini-2.5-Pro & 23.08 & 41.62~\sgain{+18.54} & 66.49~\sgain{+43.41} \\
Gemini-3-Pro   & 22.11 & 46.38~\sgain{+24.27} & 61.20~\sgain{+39.09} \\
GPT-5          & 31.09 & 60.32~\sgain{+29.23} & 65.61~\sgain{+34.52} \\
GPT-4o         & 13.46 & 13.33~\sloss{-0.13}  & 7.05~\sloss{-6.41}   \\
\midrule
\multicolumn{4}{@{}l}{\cellcolor{gray!12}\textit{Open-Source Models}} \\
Qwen3-VL-2B    & 16.19 & 11.29~\sloss{-4.90}  & 11.64~\sloss{-4.55}  \\
Qwen3-VL-8B    & 12.50 & 2.82~\sloss{-9.68}   & 13.40~\sgain{+0.90}  \\
InternVL3.5-2B & 18.11 & 32.80~\sgain{+14.69} & 25.40~\sgain{+7.29}  \\
InternVL3.5-8B & 16.67 & 12.35~\sloss{-4.32}  & 8.64~\sloss{-8.03}   \\
\bottomrule
\end{tabular}
\end{minipage}
\hfill
\begin{minipage}[t]{0.49\linewidth}
\centering
\textbf{(b) Egocentric Distance} \\[0.4em]
\begin{tabular}{@{}l ccc@{}}
\toprule
Method & Global & L-Video & L-Image \\
\midrule
\multicolumn{4}{@{}l}{\cellcolor{gray!12}\textit{Proprietary Models}} \\
Gemini-2.5-Pro & 35.40 & 46.94~\sgain{+11.54} & 53.39~\sgain{+17.99} \\
Gemini-3-Pro   & 27.39 & 48.58~\sgain{+21.19} & 55.79~\sgain{+28.40} \\
GPT-5          & 27.39 & 26.01~\sloss{-1.38}  & 31.04~\sgain{+3.65}  \\
GPT-4o         & 10.13 & 23.66~\sgain{+13.53} & 22.68~\sgain{+12.55} \\
\midrule
\multicolumn{4}{@{}l}{\cellcolor{gray!12}\textit{Open-Source Models}} \\
Qwen3-VL-2B    &  6.55 & 17.98~\sgain{+11.43} & 14.48~\sgain{+7.93}  \\
Qwen3-VL-8B    & 11.59 & 17.10~\sgain{+5.51}  & 26.12~\sgain{+14.53} \\
InternVL3.5-2B &  4.56 & 7.32~\sgain{+2.76}   & 7.92~\sgain{+3.36}   \\
InternVL3.5-8B &  8.36 & 28.42~\sgain{+20.06} & 25.79~\sgain{+17.43} \\
\bottomrule
\end{tabular}
\end{minipage}

\vspace{0.3em}

\end{table}

\paragraph{Proprietary models see locally but fail to integrate globally.}
On ED$_s$, removing the cross-frame reasoning requirement nearly triples Gemini-3-Pro's score (22.11 $\to$ 61.20) and yields gains of $+34$ to $+43$ points for all three top proprietary models. The same pattern holds, more mildly, on EDist$_s$ ($+18$ to $+28$ points for the two Gemini variants). The one anomaly is GPT-5 on EDist$_s$ Local-Video ($-1.38$), where adding a now-redundant video appears to mildly distract the model. Taken together, these results pinpoint the bottleneck: \textbf{the proprietary failure on GST-Bench is one of cross-frame spatial integration, not of single-image spatial perception.}

\paragraph{Open-source models fail at both perception and integration.}
Unlike proprietary models, open-source models do not consistently benefit from the local setting. On ED$_s$, gains range from $+7.29$ (InternVL3.5-2B) to $-8.03$ (InternVL3.5-8B); two of the four evaluated models actually \textbf{degrade} when the task is made easier. On EDist$_s$ all four models improve, but the absolute local scores remain far below proprietary counterparts — InternVL3.5-2B reaches only 25.40 on ED$_s$ Local-Image versus Gemini-2.5-Pro's 66.49. This reveals a qualitatively different bottleneck: \textbf{whereas proprietary models see locally but fail to integrate globally, open-source models fail at both stages}, with local perception itself being the more pressing limitation.

\subsection{Bridging the Gap with Targeted Training}

We further evaluate whether the deficiencies exposed by GST-Bench can be alleviated through targeted supervision. Specifically, we fine-tune Qwen3-VL-8B on GST-Train, which is generated by the same pipeline as GST-Bench with scenes disjoint from the evaluation split. To preserve general multimodal instruction-following ability, we mix GST-Train with general-purpose multimodal instruction data during supervised fine-tuning.

Table~\ref{tab:main_results} shows that this targeted fine-tuning substantially improves Qwen3-VL-8B, raising its average score from 25.89 to 53.52. The fine-tuned model surpasses all zero-shot models evaluated in this work, including proprietary systems, demonstrating that explicit supervision for global spatial reasoning can significantly improve long-horizon spatial memory and cross-view alignment. At the same time, the remaining gap to human performance indicates that targeted supervision improves, but does not fully solve, the long-horizon spatial reasoning challenge posed by GST-Bench.
\section{Conclusion}
\label{sec:Conclusion}

We introduce GST-Bench, a video-based benchmark for evaluating global spatial awareness in VLMs. By requiring models to reason from long-horizon egocentric videos, novel off-trajectory viewpoints, and explicit top-down scene representations, GST-Bench targets spatial abilities that are essential for embodied agents but underexplored in existing benchmarks. Our evaluation of 22 state-of-the-art VLMs reveals a substantial gap between current models and humans, showing that global spatial reasoning remains a major limitation even for the strongest proprietary and embodied-understanding models. Through controlled local variants and GST-Train, we further show that this gap stems largely from failures in cross-frame spatial integration and can be partially narrowed with targeted supervision. We hope GST-Bench and GST-Train will facilitate future research toward VLMs that can build, maintain, and reason over globally consistent spatial representations.

%\clearpage

\bibliographystyle{plainnat}
\bibliography{main}

%\clearpage
%\beginappendix
%\input{sections/appendix}

\end{document}